\documentclass[conference]{IEEEtran}
\IEEEoverridecommandlockouts
\usepackage{cite}
\usepackage{amsmath,amssymb,amsfonts}
\usepackage{graphicx}
\usepackage{textcomp}
\usepackage{xcolor}
\usepackage{booktabs}
\usepackage[caption=false,font=footnotesize]{subfig}
\usepackage{url}
\usepackage[hidelinks]{hyperref}
\usepackage{amssymb}
\usepackage{subcaption}
\usepackage{float} 
\usepackage[ruled,vlined]{algorithm2e}
\def\BibTeX{{\rm B\kern-.05em{\sc i\kern-.025em b}\kern-.08em
    T\kern-.1667em\lower.7ex\hbox{E}\kern-.125emX}}
\begin{document}

\title{Opportunity-Cost Option-Critic: Adaptive Switching Costs for Robust Hierarchical Reinforcement Learning}

\author{%
\makebox[\textwidth][c]{%
\begin{minipage}[t]{0.31\textwidth}
\centering
\textbf{Khaled Ammoura}\\
\textit{Department of Electrical and}\\
\textit{Computer Engineering}\\
\textit{American University of Beirut}\\
Beirut, Lebanon\\
kaa74@mail.aub.edu
\end{minipage}
\hfill
\begin{minipage}[t]{0.31\textwidth}
\centering
\textbf{Aya Kayal}\\
\textit{Department of Computer Science}\\
\textit{American University of Beirut}\\
Beirut, Lebanon\\
ak29@aub.edu.lb
\end{minipage}
\hfill
\begin{minipage}[t]{0.31\textwidth}
\centering
\textbf{Naseem Daher}\\
\textit{Department of Electrical and}\\
\textit{Computer Engineering}\\
\textit{Department of Mechanical Engineering}\\
\textit{American University of Beirut}\\
Beirut, Lebanon\\
nd38@aub.edu.lb

\end{minipage}
}%
}
\maketitle

\begin{abstract}

Deliberation costs are commonly used in hierarchical reinforcement learning (HRL) to prevent options from collapsing to primitive actions. Existing approaches rely on fixed penalties or duration-based adaptations, both of which are indirect proxies for an option's utility. This paper argues that option persistence should instead depend on the opportunity cost of abandoning the current behavior. We propose the Opportunity-Cost Option-Critic ($OC^2$) framework, which defines a state-dependent deliberation cost using the positive value gap between the active option and its best alternative. This encourages persistence when the current option is clearly preferable and allows rapid switching when a better option emerges. We show that directly learning such costs leads to a degenerate zero-cost solution and introduces an auxiliary objective that aligns the learned cost with the value-gap target. The approach is evaluated in Four Rooms and Pinball domains under stationary, goal-shift, and geometry-shift conditions. In the Pinball goal-shift task, the $OC^2$ framework achieves a 52\% first-ten-episode success rate compared with 18\% for a fixed-cost baseline. The results suggest that opportunity-cost-based persistence improves adaptation in non-stationary environments.

\end{abstract}


\section{Introduction}

% Navigation in dynamically changing environment needs a controller that retains good behavior while abandoning those that give unexpected outcomes. A path-following policy would work perfectly in a stationary environment, yet if a goal or geometry shift happened to the environment it would struggle. This creates a temporal decision problem: how long should a behavior persist, and when is the right time for a higher-level decision maker to take over.

Many autonomous systems operate in environments that evolve over time, requiring controllers to preserve effective behaviors while abandoning those that are no longer beneficial. Examples include navigation, robotic manipulation, and multi-stage control tasks, where changes in objectives or operating conditions may render a previously effective behavior suboptimal. Such situations create a temporal decision problem: how long should a behavior persist, and when should a higher-level decision maker switch to an alternative strategy?

Hierarchical reinforcement learning (HRL) addresses these inquiries by adopting the temporal abstraction concept by extending multiple primitive steps \cite{barto2003}. The \textit{options} framework, proposed by \cite{sutton1999}, represents such behaviors using an initiation set, internal policy, and a termination function. In \cite{bacon2017}, the Policy Gradient Theorem was generalized in order to learn the internal policies and termination functions of the options from the task reward. However, learning to terminate does not guarantee temporal abstraction, since the options could learn to collapse to primitivity, which motivates the introduction of a stabilizing parameter that penalizes overly frequent option terminations and switches.

Deliberation costs penalize option termination and thereby regulate the trade-off between maintaining the current option and selecting an alternative. The fixed deliberation cost proposed in \cite{harb2018} guarantees that behavior, but its effect depends on the value functions that the environment learns, and the influence on termination depends on the magnitude of the option advantage relative to the chosen fixed cost. A penalty that puts pressure on termination in one setting could be neglected with respect to another. The flexible deliberation cost proposed in \cite{leal2025} is a duration-dependent cost that proposes an adaptive alternative to that of \cite{harb2018}, however, the option's duration does not by itself guarantee its optimality at the current state. Moreover, the policy updates could favor a prolonged option that would eventually dominate, causing overcommitment.

% In our work, we investigate the possibility for the deliberation cost to be value-dependent, in which, it is derived from the positive gap between the current option's value relative to the distinct best alternative. Thus, this cost reflects the opportunity cost driven by the abandoning of a currently favorable behavior. In case another option become a better alternative, the cost is zero. Following this mechanism, it removes the fixed cost manual tuning, and reduces unnecessary persistence when a better alternative emerges.

In this work, we argue that option persistence should be governed by the opportunity cost of abandoning the current behavior. To this end, we define a value-dependent deliberation cost derived from the positive gap between the active option and its best distinct alternative.
% We, therefore, have three contributions. First, we introduce a $OC^2$ deliberation cost, in which, its intended advantage is its penalty adaptation to value estimates rather than requiring a manually selected fixed cost. Second, we give a consistent cost-augmented formulation and how we came up with the auxiliary cost fitting, including a short negative degenerate result produced by direct cost-learning. Third, we evaluated this mechanism against the other configurations across ten randoms seeds in a navigation task in 3 configurations: stationary, goal-shifted, and geometry-shifted configuration. Together, these contributions examine cost-adaptive persistence within the establish Option-Critic framework.
Thus, the main contributions of this work are as follows:
\begin{enumerate}
    \item We introduce an opportunity-cost view of option persistence in hierarchical reinforcement learning, arguing that persistence should be governed by the relative value of continuing the current behavior versus switching to an alternative.
    \item We instantiate this principle through the $OC^2$ framework, a value-gap-based extension of the Option-Critic architecture that replaces fixed and duration-dependent deliberation costs with a state- and option-dependent opportunity-cost formulation.
    \item We demonstrate that opportunity-cost-based persistence improves adaptation to task changes while preserving the benefits of temporal abstraction.
\end{enumerate}

The remainder of this paper is organized as follows. Section \ref{background_related_work} reviews the background and related work. Section \ref{OC2_framework} presents the proposed $OC^2$ framework. Section \ref{experimental_eval} describes the experimental setup and results. Section \ref{Discussion and Limitations} discusses the persistence signal in our proposed framework in addition to the limitations that this framework has.Finally, Section \ref{conclusion} concludes the paper and outlines the directions for future work

\section{Background and Related Work}
\label{background_related_work}
\subsection{Preliminaries}

Consider a Markov Decision Process, which is a 5-tuple space containing each of: state space represented as $S$; action space represented as $A$; transition kernel $P(s'|s,a)$, where $P:S\times A\times S \rightarrow [0,1]$; expected reward $r(s,a)$, where $r:S\times A \rightarrow \mathbb{R}$; and a discount factor $\gamma \in (0,1)$. An option, defined as a temporally-extended action, is denoted as $\omega \in \Omega$, where $\Omega$ is the option space comprising $(I_{\omega},\pi_{\omega,\theta},\beta_{\omega,\vartheta})$, where $I_{\omega}\subseteq S$ is the initiation set of the option $\omega$, $\pi_{\omega,\theta}:S\times A \rightarrow [0,1]$ is the option's internal policy, and $\beta_{\omega,\vartheta}:S \rightarrow [0,1]$ is its termination function  \cite{sutton1999}. We make the same assumption as in \cite{bacon2017, harb2018} such that the initiation set of every option is the entire state space ($I=S$); that is, any option can be initiated from any state. Upon termination, decision-making returns to the policy-over-options, denoted as $\pi_{\Omega}(\omega|s)$, where $\pi_{\Omega}:S\times\Omega\rightarrow[0,1]$, which selects the next option.

The Option-Critic architecture in \cite{bacon2017} generalizes the Policy Gradient framework of \cite{sutton2000}, in which it derives policy-gradient updates for internal policies and termination functions. Its termination signal compares the current option value with the value of reselection. Although, the Option-Critic architecture adopts \textit{options} to achieve temporal abstraction, it does not guarantee that the learned behaviors remain temporally extended. In practice, training may converge to a degenerate solution in which options terminate at nearly every time step, effectively collapsing to primitive actions. 
% not converging to primitivity, in which, the options learned terminate each timestep, or in that proximity.

\subsection{Related Work}

To encourage option persistence, a deliberation cost was introduced into the Option-Critic architecture to penalize option termination and discourage unnecessary switching between options \cite{harb2018}. This approach was later extended in \cite{klissarov2017} by integrating it with proximal policy optimization (PPO) for continuous-action tasks. However, the deliberation cost proposed in \cite{harb2018} requires careful tuning to exert the intended influence on the value estimates. If set too low, it may fail to discourage excessive switching; if set too high, it may prevent the agent from adapting when alternative behaviors become preferable. More recently, \cite{leal2025} extended the fixed-deliberation cost of \cite{harb2018} to a flexible duration-dependent one.
% In which, the proposed adaptive cost is dependent on the option's duration. If the duration was very long, then the cost tends to zero, otherwise it would have an adaptive pressure on the value estimates.
In this formulation, the deliberation penalty decreases with option duration, gradually approaching zero for long-lived options while exerting stronger pressure on value estimates for short-lived ones.
However, \cite{leal2025} assumes that option duration is an appropriate measure of persistence. In contrast, our proposed approach defines deliberation costs using the value gap between the active option and its best alternative, directly capturing the opportunity cost of switching. As a result, duration-based costs may lead to option overcommitment, potentially favoring persistence even when switching would be optimal.

In this work, we focus specifically on the mechanism governing option persistence while retaining the structure of the Option-Critic architecture. In particular, we aim to address the hyperparameter-tuning requirement of \cite{harb2018} and the reliance on option duration as a proxy for persistence in \cite{leal2025}. To this end, we leverage the value estimates induced by the environment dynamics to adapt the deliberation cost, allowing persistence pressure to emerge from the relative value of competing options rather than from a fixed parameter or elapsed duration. While this work focuses on deliberation costs, other studies have addressed different aspects of temporal abstraction. For example, the Termination Critic in \cite{harutyunyan2019} learns option termination functions using an information-theoretic objective. Similarly,\cite{khetarpal2020} extends the notion of initiation sets by learning when options should be available, rather than assuming predefined option availability.
% Note that, in our work, which is built on the previous Option-Critic work, assumes that all options can start from any state, meaning $I=S$. All these methods, illustrates that an option is not only duration-variant concept.
Note that our work, like the original Option-Critic framework, assumes that every option is available in every state ($I=S$). While temporal abstraction involves multiple mechanisms, this work focuses specifically on option persistence.

\section{Opportunity-Cost Option-Critic ($OC^2$)}
\label{OC2_framework}

The proposed framework builds upon the call-and-return execution model, value functions, and actor updates of the Option-Critic architecture \cite{bacon2017}, while adopting the deliberation cost extension of \cite{harb2018}. The key difference lies in how the deliberation cost is defined and adapted. In \cite{harb2018}, a scalar coefficient $\eta$ is used as a fixed-cost coefficient. In contrast, \cite{leal2025} introduces an adaptive duration-dependent deliberation cost. Our approach adopts the deliberation-cost framework of \cite{harb2018} and the adaptive-cost extension of \cite{leal2025}, in which, it learns state-and-option-dependent cost $C_{\phi}(s,\omega)$ that eliminates the need for a fixed penalty and avoids assuming that optimal persistence is determined solely by option duration.

\subsection{Cost-augmented evaluation and timing}

At the primitive step $t$, let $s_t$ be the state and $\omega_t$ be the current option. Let $f_\phi(s,\omega)$ denote the raw cost prediction. We enforce nonnegativity by defining
\[
    C_\phi(s,\omega)=\max\{0,f_\phi(s,\omega)\}.
\]

On a nonterminal arrival, let \(\beta_{t+1}=\beta_{\omega_t,\vartheta}(s_{t+1})\) be the probability that the active option terminates. For the deliberation-cost calculation, set \(\beta_{t+1}=0\) on terminal transitions. With discount factor $\gamma$, the expected cost-augmented return is

\begin{equation}
    J^C=\mathbb{E}\!\left[
    \sum_{t\geq 0}\gamma^t
    \bigl(r_{t+1}-\beta_{t+1}C_\phi(s_{t+1},\omega_t)\bigr)
    \right],
    \label{equation:1}
\end{equation}
where the superscript $C$ denotes values computed with the deliberation cost included.

The option value $Q_\Omega^C(s,\omega)$ averages the action value $Q_U^C(s,\omega,a)$ over the intra-option policy $\pi_{\omega,\theta}(a\mid s)$. Here, $\theta$ parameterizes the policy, and $Q_U^C(s,\omega,a)$ is the expected cost-augmented return from taking primitive action $a$ in state $s$ with option $\omega$ active and then following the agent's policies:

\begin{align}
    Q_\Omega^C(s,\omega)
    &= \sum_a \pi_{\omega,\theta}(a\mid s)\,Q_U^C(s,\omega,a).
\end{align}

The policy over options, $\pi_\Omega(\omega\mid s)$, gives the
probability of selecting $\omega$ when an option is chosen in $s$.
The corresponding reselection value is
\begin{align}
    V_\Omega^C(s)
    &= \sum_{\omega\in\Omega}\pi_\Omega(\omega\mid s)\,
       Q_\Omega^C(s,\omega).
\end{align}

At a nonterminal next state $s'$, $\beta_{\omega,\vartheta}(s')$
is the probability of terminating the active option $\omega$,
with $\vartheta$ denoting the termination parameters.
The arrival value $U^C(\omega,s')$ averages the future return
under continuation and reselection. The immediate deliberation
cost is accounted for separately in the Bellman recursion as follows:
\begin{align}
    U^C(\omega,s')
    &= (1-\beta_{\omega,\vartheta}(s'))Q_\Omega^C(s',\omega)
       \notag\\
    &\quad +\beta_{\omega,\vartheta}(s')V_\Omega^C(s').
    \label{eq:4.1}
\end{align}

The option advantage measures how much more valuable it is to
continue $\omega$ in $s$ than to draw an option from $\pi_\Omega$:
\begin{align}
    A_\Omega^C(s,\omega)
    &= Q_\Omega^C(s,\omega)-V_\Omega^C(s).
\end{align}

\noindent To simplify notation, parameter subscripts are omitted where unambiguous. 

Let $r(s,a)$ be the expected immediate reward, and let
$\mathbb{E}_{s'\mid s,a}$ average over the next-state distribution.
Writing $\beta_\omega$ for $\beta_{\omega,\vartheta}$, the Bellman
recursion consistent with \eqref{equation:1} is given by:
\begin{align}
    Q_U^C(s,\omega,a)
    &= r(s,a)+\mathbb{E}_{s'\mid s,a}\!\left[
       -\beta_\omega(s')C_\phi(s',\omega) \right. \notag\\
    &\hspace{5em}\left. +\gamma U^C(\omega,s')\right].
    \label{bellman}
\end{align}
The first term inside the expectation is the expected immediate
cost, while the second is the discounted arrival value. For terminal
transitions, both terms are zero.

Under this timing convention, setting $C_\phi(s',\omega)=\gamma\eta$
recovers the immediate cost term $-\gamma\eta\beta_\omega(s')$
in the fixed-cost Bellman recursion of \cite{harb2018}, showing that the proposed formulation reduces to the fixed-cost case when the learned cost is constant.

\subsection{A value-gap target for persistence}

The opportunity-cost target is the positive part of the value gap
between the current option $\omega$ and its best distinct alternative:

\begin{equation}
    y_C(s,\omega)
    = \max\!\left\{0,\,
        Q_\Omega^C(s,\omega)
        - \max_{\bar{\omega}\ne\omega} Q_\Omega^C(s,\bar{\omega})
      \right\}.
    \label{eqaution: 9}
\end{equation}

When the expected value of the chosen alternative exceeds that of all other possible choices, $y_C > 0$. When another possible choice has an equal or better expected value than the chosen alternative, $y_C = 0$. Upon termination, the policy over options $\pi_\Omega(\cdot\mid s)$ may reselect $\omega$ or a non-maximizing alternative. The expected value after reselection is therefore $V_\Omega^C(s)$, and the corresponding value difference is $A_\Omega^C(s,\omega)$.

% The present formulation adds value to the termination benefits of using the standard formulations. 
The proposed formulation provides additional insights into termination behavior by linking persistence to the opportunity cost of switching. Under a greedy policy-over-options and an exact cost function, two observations follow: First, the cost encourages persistence when the active option uniquely maximizes the objective; i.e., $A_\Omega^C = 0$ and $y_C(s,\omega) > 0$. Second, the formulation favors termination when the active option becomes less advantageous to an available alternative; i.e., $A_\Omega^C < 0$ and $y_C(s,\omega) = 0$. Under an $\epsilon$-soft policy, the advantage is evaluated relative to the expected value of the option distribution, whereas the target cost remains based on the single alternative option offering the largest  value improvement. The above observations assume a fixed critic and an exact target cost. In practice, approximation errors and imperfect target tracking may cause deviations from the actual behavior of the system.

\subsection{Why the cost uses an auxiliary objective}

Direct maximization of (\ref{equation:1}) with respect to the cost fails to identify a meaningful deliberation penalty. We hold the intra-option policies, the termination functions, and the policy over options fixed when differentiating with respect to $\phi$. The trajectory distribution is therefore independent of $\phi$. Let $J^0$ denote the reward-only return. Then,

\begin{align}
    J^C=J^0-\mathbb{E}[\sum_{t \geq 0}\gamma^{t}\beta_{t+1}C_{\phi}(s_{t+1},\omega_t)] \leq J^0,
\end{align}
indicating $J^0$ as an upper bound. More explicitly, under differentiable costs and policies,

\begin{align}
    \nabla_{\phi} J^C=-\mathbb{E}[\sum_{t \geq 0}\gamma^{t}\beta_{t+1}\nabla_{\phi}C_{\phi}(s_{t+1},\omega_t)],
\end{align}
which motivates the introduction of a separate objective. Therefore, we fit $C_{\phi}$ to the detached target in (\ref{eqaution: 9}). Let $\nu$ denote the distribution of state-option pairs used for this regression. The stop-gradient operator $\operatorname{sg}$ treats $y_C$ as constant when differentiating the loss with respect to $\phi$:
\begin{equation}
    \mathcal{L}_C(\phi)
    = \tfrac12 \mathbb{E}_{(s,\omega)\sim\nu}\!\left[
        \bigl(C_\phi(s,\omega)
        -\operatorname{sg}[y_C(s,\omega)]\bigr)^2
    \right].
    \label{eq:costloss}
\end{equation}

The critic and the learned cost are coupled across updates. The Bellman target in \eqref{bellman} uses the current cost $C_\phi$, while the cost target in \eqref{eqaution: 9} is computed from the current cost-augmented option values. The stop-gradient operator $\operatorname{sg}$ holds the target fixed when differentiating $\mathcal{L}_C$ with respect to $\phi$. As the critic is updated, its option values can change, causing the cost regression to track a moving target. The convergence properties of these coupled updates remain an open question for this formulation.

\subsection{Actor updates and the termination signal}

For fixed cost parameters $\phi$, termination parameters $\vartheta$, and policy over options $\pi_\Omega$, the intra-option policy gradient keeps the Option-Critic form with cost-augmented action values \cite{bacon2017}:

\begin{equation}
 \nabla_\theta J^C=\sum_{s,\omega}d_\gamma(s,\omega)
 \sum_a\nabla_\theta\pi_{\omega,\theta}(a\mid s)Q_U^C(s,\omega,a),
 \label{eq:policygradient}
\end{equation}
where $d_\gamma(s,\omega)=\sum_{t\geq0}\gamma^t
\Pr(s_t=s,\omega_t=\omega)$.

For the termination gradient, we hold $\theta$, $\phi$, and $\pi_\Omega$ fixed and evaluate the branch at the nonterminal arrival state $s'=s_{t+1}$, before deciding whether to terminate $\omega_t$.
For termination, the local branch value in~\eqref{bellman} is:
\begin{equation}
\begin{aligned}
 &-\beta_\omega(s')C_\phi(s',\omega)\\
 &\quad+\gamma\bigl[(1-\beta_\omega(s'))Q_\Omega^C(s',\omega)
                  +\beta_\omega(s')V_\Omega^C(s')\bigr].
\end{aligned}
\end{equation}
Holding $C_\phi(s',\omega)$, $Q_\Omega^C(s',\omega)$, and
$V_\Omega^C(s')$ fixed locally, the derivative with respect to
$\beta_\omega(s')$ is
\begin{equation}
\begin{aligned}
 &-C_\phi(s',\omega)
   +\gamma\bigl[V_\Omega^C(s')-Q_\Omega^C(s',\omega)\bigr]\\
 &\quad=-\bigl[C_\phi(s',\omega)+\gamma A_\Omega^C(s',\omega)\bigr].
\end{aligned}
\label{eq:localtermination}
\end{equation}
Unrolling the derivative through the augmented state-option process yields
\begin{equation}
 \nabla_\vartheta J^C
 = -\mathbb{E}\!\left[
 \sum_{t\geq0}\gamma^t
 \nabla_\vartheta\beta_{\omega_t,\vartheta}(s_{t+1})
 H_C(s_{t+1},\omega_t)
 \right],
 \label{eq:terminationgradient}
\end{equation}
with $H_C(s,\omega)=C_\phi(s,\omega)+\gamma A_\Omega^C(s,\omega)$.

Under our timing convention, setting $C_\phi=\gamma\eta$ yields $H_C=\gamma(A_\Omega^C+\eta)$, thereby recovering the fixed termination margin that was derived in \cite{harb2018}. In our formulation, the cost within this margin depends on both the arrival state and active option and is learned from the value-gap target in (\ref{eqaution: 9}).

\subsection{Algorithm and control implementation}

Algorithm \ref{alg:ococ} summarizes the proposed update procedure. The first sampled transition provides the critic target. The policy is then updated using the cost-augmented value estimate, followed by an update of the termination function through $H_C$. Finally, the cost is fitted to the detached value-gap target. Let $\psi$ denote the parameters of the critic $Q_{U,\psi}^{C}(s,\omega,a)$, which approximates the cost-augmented action value $Q_{U}^{C}(s,\omega,a)$.

\begin{algorithm}[t]
\caption{Opportunity-Cost Option-Critic}
\label{alg:ococ}
\small
\DontPrintSemicolon

Initialize the critic parameters \(\psi\), intra-option policy parameters \(\theta\), termination parameters \(\vartheta\), and cost parameters \(\phi\);

Set the corresponding learning rates \(\alpha_\psi,\alpha_\theta,\alpha_\vartheta,\alpha_\phi\);

\ForEach{episode}{
    $s\gets s_0,\qquad \omega\sim\pi_\Omega(\cdot\mid s)$\;

    \Repeat{$d=1$}{
        Choose $a$ according to $\pi_{\omega,\theta}(a|s)$;\\
        Take action \(a\) in \(s\) and observe the next state \(s'\), reward \(r\), and terminal flag \(d\), where \(d=1\) if the transition ends the episode and \(d=0\) otherwise;

        \textbf{1. Option evaluation}\;
        $z\gets r$\;
        \If{$d=0$}{
            $c\gets C_\phi(s',\omega)$\;
            $z\gets r-\beta_{\omega,\vartheta}(s')c+\gamma U^C(\omega,s')$
                \hfill\eqref{bellman}\;
            $\begin{aligned}[t]
                y\gets\max\bigl\{0,\;&Q_\Omega^C(s',\omega)\\
                &-\max_{j\ne\omega}Q_\Omega^C(s',j)\bigr\}
            \end{aligned}$
                \hfill\eqref{eqaution: 9}\;
            $h\gets c+\gamma A_\Omega^C(s',\omega)$\;
        }
        $\delta\gets z-Q_{U,\psi}^C(s,\omega,a)$\;
        $\psi\gets\psi+\alpha_\psi\delta\,
          \nabla_\psi Q_{U,\psi}^C(s,\omega,a)$\;

        \textbf{2. Option improvement and cost learning}\;
        $\theta\gets\theta+\alpha_\theta Q_{U,\psi}^C(s,\omega,a)
          \nabla_\theta\log\pi_{\omega,\theta}(a\mid s)$\;
        \If{$d=0$}{
            $\vartheta\gets\vartheta-\alpha_\vartheta h\,
                \nabla_\vartheta\beta_{\omega,\vartheta}(s')$
                \hfill\eqref{eq:terminationgradient}\;
            $\phi\gets\phi+\alpha_\phi(y-c)\nabla_\phi c$
                \hfill\eqref{eq:costloss}\;
            \If{$\omega$ terminates at $s'$ under the pre-update
                $\beta_{\omega,\vartheta}$}{
                $\omega\sim\pi_\Omega(\cdot\mid s')$\;
            }
        }
        $s\gets s'$\;
    }
}
\end{algorithm}

\begin{table*}[t]
\centering
\caption{Four Rooms evaluation after 5,000 training episodes. All methods complete in 100\% of evaluation episodes. Values are mean $\pm$ sample standard deviation across ten seed-level averages (each seed level averaged across ten evaluation episodes), with 100 evaluation episodes per seed. Steps number primitive transitions per episode. Discounted return is the discounted sum of rewards per episode ($\gamma=0.99$). Terminations number option-ending decisions per episode (including option terminations followed by re-selection of the same option). Switches number transitions to a different option per episode. Duration is the mean number of primitive steps per execution of an option.}
\label{tab:fourrooms}
\small\setlength{\tabcolsep}{15pt}
\begin{tabular}{lrrrrr}
\toprule
Method & Steps & Discounted return & Terminations & Switches & Duration\\
\midrule
Vanilla OC & $26.56\pm9.06$ & $0.794\pm0.053$ & $17.11\pm7.31$ & $7.41\pm4.51$ & $1.46\pm0.06$\\
Fixed, $\eta=0.005$ & $32.99\pm20.55$ & $0.777\pm0.087$ & $18.45\pm13.26$ & $10.30\pm9.39$ & $1.73\pm0.18$\\
Fixed, $\eta=0.01$ & $30.23\pm8.55$ & $0.777\pm0.050$ & $11.47\pm3.57$ & $5.15\pm2.49$ & $2.33\pm0.50$\\
Flexible cost & $68.67\pm36.03$ & $0.623\pm0.102$ & $1.55\pm0.58$ & $1.02\pm0.47$ & $27.28\pm16.65$\\
 $OC^2$ & $\mathbf{19.11\pm4.89}$ & $\mathbf{0.843\pm0.032}$ & $6.72\pm1.76$ & $4.95\pm1.64$ & $2.99\pm1.45$\\
\bottomrule
\end{tabular}
\end{table*}

\section{Experimental Evaluation}
\label{experimental_eval}

\subsection{Questions, protocol, and reporting}

This work addresses three limitations of existing deliberation-cost mechanisms through the following questions: (i) Does opportunity-cost-based persistence improve navigation performance beyond merely increasing option duration? (ii) Can the proposed approach successfully and rapidly adapt to changes in goal location? (iii) Does the proposed approach maintain its performance under changes in environment geometry? Results are reported across ten random seeds for each method under comparison. 

\begin{figure}
    \centering
    \includegraphics[width=0.5\linewidth]{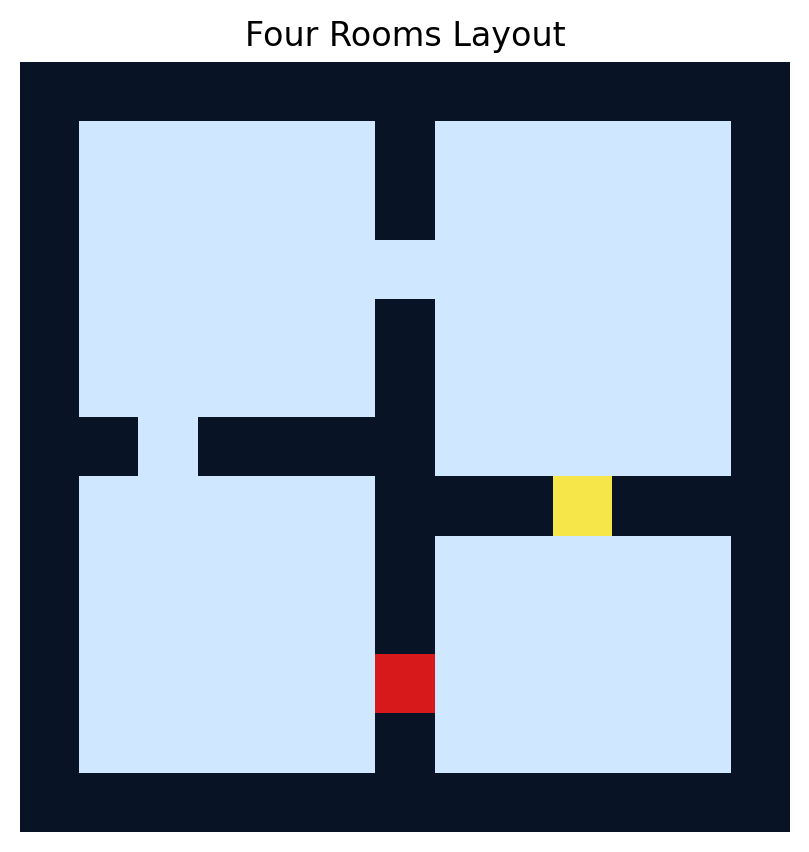}
    \caption{Four Rooms environment where the starting point is colored red and the goal is colored yellow. The choice of the starting point and goal is random.}
    \label{fig:fourrooms}
\end{figure}

Two environments are used in this study for different purposes. The Four Rooms environment (Fig.~\ref{fig:fourrooms}) is used to evaluate the learned policies after the training period. In contrast, the Pinball environment is considered to assess the online training performance. In fact, the Pinball statistics' success rates reflect both the learning process and adaptation during a specified time frame, as opposed to a separate assessment of the trained policy itself. For each of the seeds, the average value and standard deviation of those seeds' summary data were calculated. The standard deviation reflects how much variability existed among the different seeds (i.e., runs) but does not represent a confidence interval. The conclusions drawn from this study are descriptive and should be interpreted in light of the uncertainty and replication challenges associated with reinforcement learning evaluation~\cite{henderson2018,agarwal2021}.

The stationary learning period of the Pinball task takes place from episodes 1-250. From episode 251 to 500, two additional learning tasks occur: one in which the target is relocated while the environment geometry remains unchanged, and another in which the obstacle geometry is modified while maintaining the original target location. The recovery rate is defined as the number of post-shift episodes until the first 20-episode window achieves $\ge90\%$ success rate. The recovery rate can vary depending on the definition of the recovery window, therefore, the number of seeds that reach the recovery threshold must be taken into account when interpreting these results. 

In both environments and during all of the considered scenarios, for fairness purposes, we report results for the fixed-cost baseline using the $\eta$ values that yielded the strongest overall performance across the evaluated environments.

\subsection{Four Rooms: persistence and navigation}

The Four Rooms environment is a discrete-state, discrete-action navigation task that has four rooms with doors leading into each other. An agent can move around in these rooms using the available doors to try finding a way to get to a specific target point. The reward is +1 at the goal and 0 otherwise. We evaluated five variants of the Option-Critic framework: standard Option-Critic~\cite{bacon2017}; fixed cost Option-Critic~\cite{harb2018} with costs of $\eta \in \{0.005, 0.01\}$; the flexible-cost method~\cite{leal2025}; and $OC^2$. All five methods were trained for 5,000 episodes and then tested on 100 random episodes from each seed. For all of the methods, the same general settings were used: four options; $\gamma = 0.99$; learning rate of $0.5$ for the critic; learning rate of $0.25$ for the policy and termination functions; softmax temperature of $0.01$; and exploration rate of $\epsilon_\Omega = 0.01$ during training.

\begin{figure*}[!t]
    \centering
    \begin{minipage}[t]{0.32\textwidth}
        \centering
        \includegraphics[width=\linewidth]{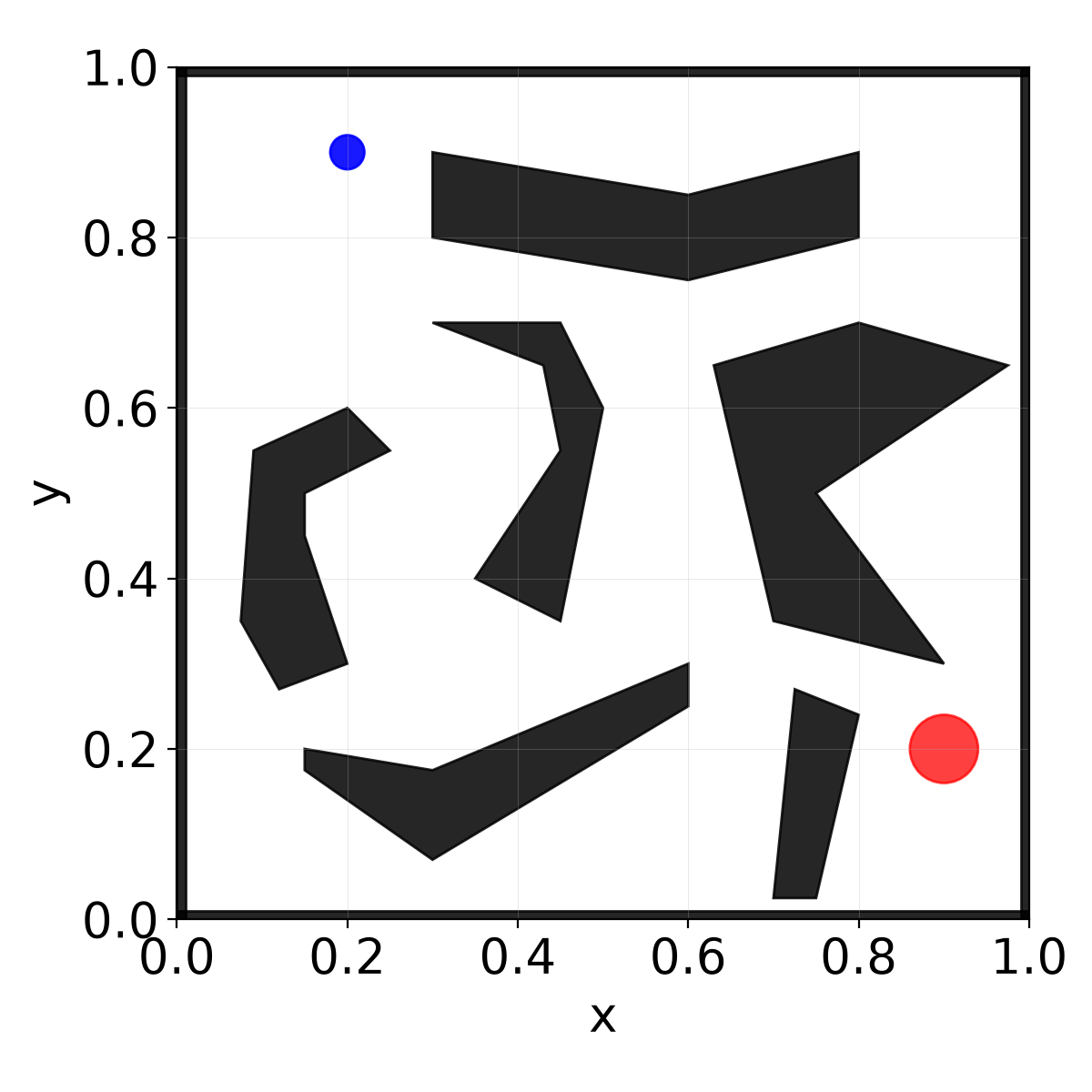}\par
        {\small (a) Original Environment\par}
    \end{minipage}\hfill
    \begin{minipage}[t]{0.32\textwidth}
        \centering
        \includegraphics[width=\linewidth]{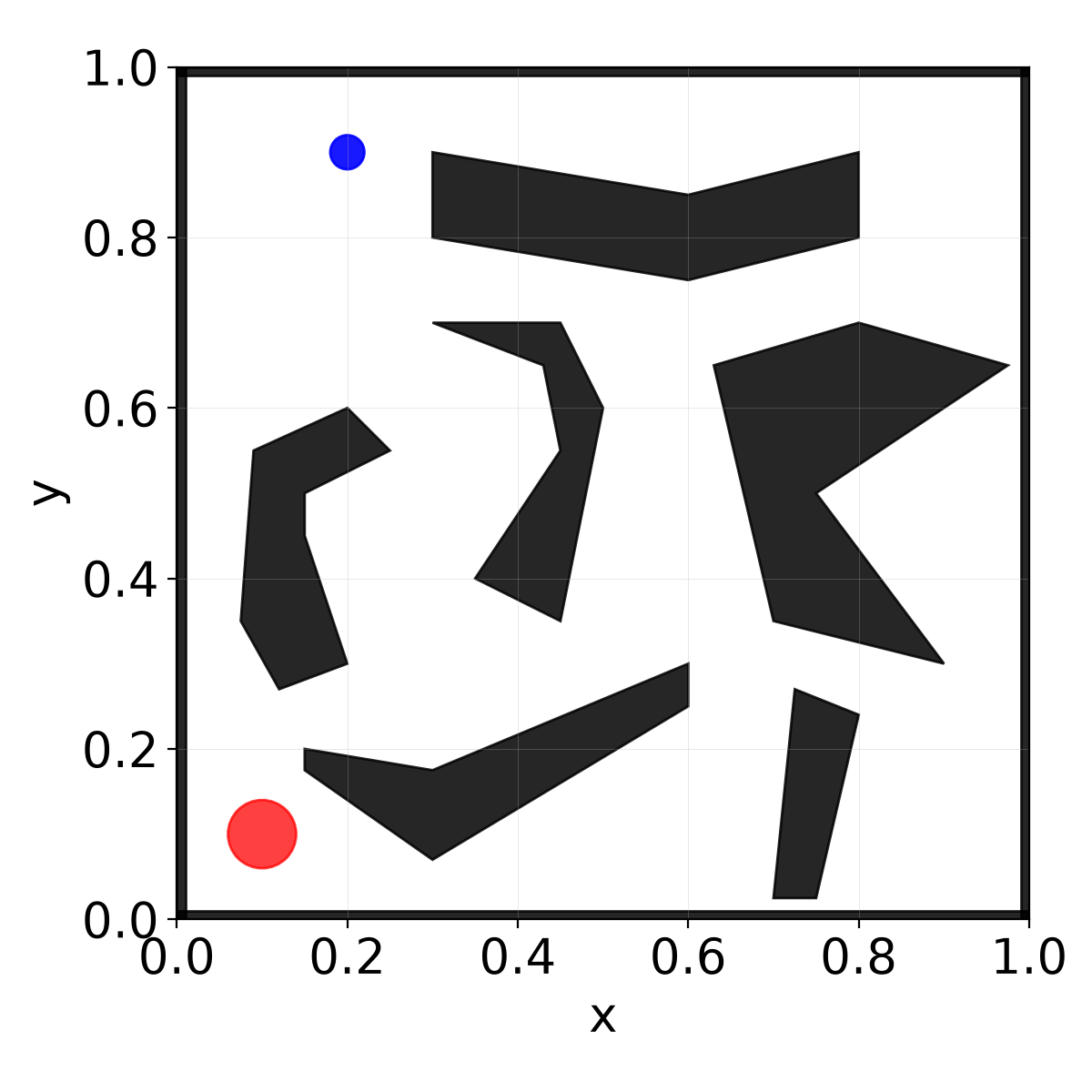}\par
        {\small (b) Relocated-Goal Environment\par}
    \end{minipage}\hfill
    \begin{minipage}[t]{0.32\textwidth}
        \centering
        \includegraphics[width=\linewidth]{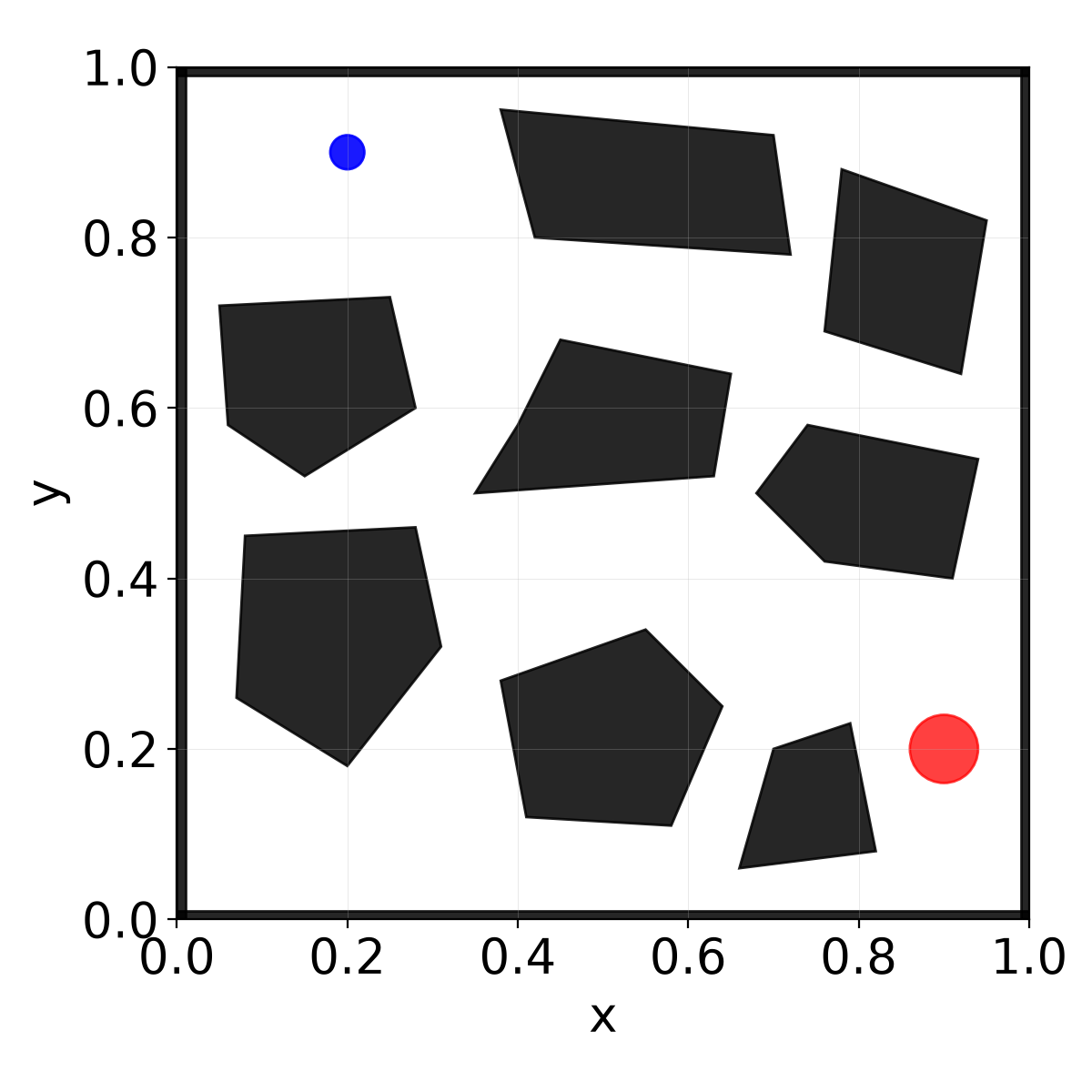}\par
        {\small (c) Altered-Geometry Environment\par}
    \end{minipage}
    \caption{Pinball environment layouts: a) original environment; b) relocated-goal environment preserves the obstacles shape; altered-geometry environment preserves the start and goal.}
    \label{fig:layouts}
\end{figure*}

Table~\ref{tab:fourrooms} demonstrates that $OC^2$ had the smallest average length of trajectory and greatest average discounted returns when compared to the other approaches evaluated. The average trajectory for $OC^2$ was approximately 28\% shorter than that of vanilla OC. While the flexible-cost baseline resulted in significantly longer option durations, it also resulted in substantially longer trajectories, indicating that persistence is not an adequate metric for evaluating option performance. $OC^2$ demonstrated moderate option duration, while at the same time having fewer option terminations and identity switches than vanilla OC. These findings support the concept of beneficial persistence in this domain; however, they do not provide evidence that the learned options are semantically specialized.

\begin{table*}[t]
\centering
\caption{Pinball play on original layout in training episodes 1--250. Success is the percentage of episodes that reach the goal. Undiscounted return is the sum of per-episode rewards, steps per episode is the number of primitive transitions. Mean $\pm$ sample standard deviation over ten seed-level summaries.}
\label{tab:pinball_stationary}

\small
\setlength{\tabcolsep}{6pt}
\begin{tabular*}{\textwidth}{@{\extracolsep{\fill}}lrrrr@{}}
\toprule
Metric
& Fixed $\gamma=.99$
& Fixed $\gamma=.999$
& Flexible $\gamma=.999$
& $OC^2$ $\gamma=.999$ \\
\midrule
Success (\%)
& $82.60\pm36.46$ & $92.56\pm10.18$
& $95.28\pm3.64$ & $99.20\pm0.38$ \\
Undiscounted return
& $4615.6\pm7289.8$ & $4491.9\pm3425.8$
& $4521.0\pm2384.3$ & $6049.6\pm963.8$ \\
Steps per episode
& $2111.2\pm3570.1$ & $1457.7\pm1144.1$
& $1187.6\pm544.2$ & $865.9\pm217.3$ \\
\bottomrule
\end{tabular*}
\end{table*}

\begin{table*}
\centering
\caption{Pinball performance in the 250 post-change training episodes following a goal or obstacle change. First-10 success is the percentage of the first ten post-change episodes that attained the goal. Success is the percentage across 250 post-change episodes. Undiscounted return is the sum of per-episode rewards; episode length is the number of primitive transitions in an episode, both averaged across the 250 post-change episodes. Values are mean $\pm$ sample standard deviation across ten seed-level summaries. Recovered seeds is the number of ten seeds in which the agent achieved at least 90\% success for any window of 20 consecutive post-change episodes.}
\label{tab:adaptation}
\small\setlength{\tabcolsep}{13pt}
\begin{tabular}{llrrrr}
\toprule
Change & Metric
& Fixed $\gamma=.99$
& Fixed $\gamma=.999$
& Flexible $\gamma=.999$
& $OC^2$ $\gamma=.999$ \\
\midrule
Goal
& First-10 success (\%)
& $18.0\pm24.4$ & $18.0\pm38.2$
& $15.0\pm26.8$ & $52.0\pm27.4$ \\
& Success (\%)
& $73.92\pm39.65$ & $87.80\pm6.32$
& $89.28\pm10.99$ & $96.76\pm3.66$ \\
& Undiscounted Return
& $2429.9\pm7438.5$ & $2635.8\pm3167.0$
& $3093.2\pm4160.8$ & $7324.7\pm1226.4$ \\
& Episode length
& $2757.4\pm3890.7$ & $1501.3\pm681.4$
& $1403.1\pm1069.7$ & $561.2\pm352.0$ \\
& Recovered seeds
& $8/10$ & $10/10$ & $10/10$ & $10/10$ \\
\midrule
Geometry
& First-10 success (\%)
& $81.0\pm40.1$ & $87.0\pm30.9$
& $74.0\pm31.7$ & $100.0\pm0.0$ \\
& Success (\%)
& $80.04\pm42.08$ & $98.80\pm2.32$
& $96.12\pm7.29$ & $99.92\pm0.17$ \\
& Undiscounted Return
& $5372.3\pm8459.5$ & $8386.7\pm1292.7$
& $6806.8\pm2261.7$ & $9018.4\pm526.8$ \\
& Episode Length
& $2096.8\pm4157.1$ & $373.8\pm272.0$
& $759.7\pm758.6$ & $198.3\pm102.8$ \\
& Recovered seeds
& $8/10$ & $10/10$ & $10/10$ & $10/10$ \\
\bottomrule
\end{tabular}
\end{table*}

\subsection{Pinball: stationary performance}

The Pinball Environment is a continuous-state, continuous-control navigation task in which a ball traverses an environment containing polygon-shaped obstacles. The state consists of the ball's position and velocity \cite{konidaris2009}. Compared with grid-based navigation tasks (e.g., Four Room), the pinball exhibits more complex dynamics due to inertia and collisions. Every thrust action results in a penalty of 5 and every non-action results in a 1 penalty. If the agent gets to the goal, +10000 reward is given. Three environment layouts are considered, as shown in Fig.~\ref{fig:layouts}. All Pinball algorithms tested in this paper utilize four options, a third-order Fourier basis~\cite{konidaris2011}, and a maximum number of steps allowed per episode of 10,000 steps. For fairness and best individual performances, the fixed-cost baseline utilizes $\eta=0.02$ with $\gamma\in\{0.99,0.999\}$, whereas the flexible-cost baseline and $OC^2$ both utilize $\gamma=0.999$. Therefore, the primary matched-discount comparison for the fixed-cost baseline is the $\gamma=0.999$ fixed-cost configuration. Moreover, the learning rate of the critic is 0.01, and 0.001 for the policy and termination functions. Note that we have emitted benchmarking against \cite{bacon2017} since the purpose of these experiments is to compare the effect of the proposed deliberation cost against the other deliberation cost models.

Within the first 250 training episodes, as shown in Table~\ref{tab:pinball_stationary}, $OC^2$ had a mean success of 99.20\%, compared to 95.28\% for the flexible-cost baseline and 92.56\% for the fixed-cost baseline at $\gamma=0.999$. Also, $OC^2$ took fewer steps to complete each episode on average (865.9 vs 1187.6 and 1457.7), which supports consistent learning on the original layout.

\subsection{Pinball: response to goal relocation}

\begin{figure*}[!t]
    \centering
    \subfloat[Relocated Goal]{%
        \includegraphics[width=0.48\textwidth]{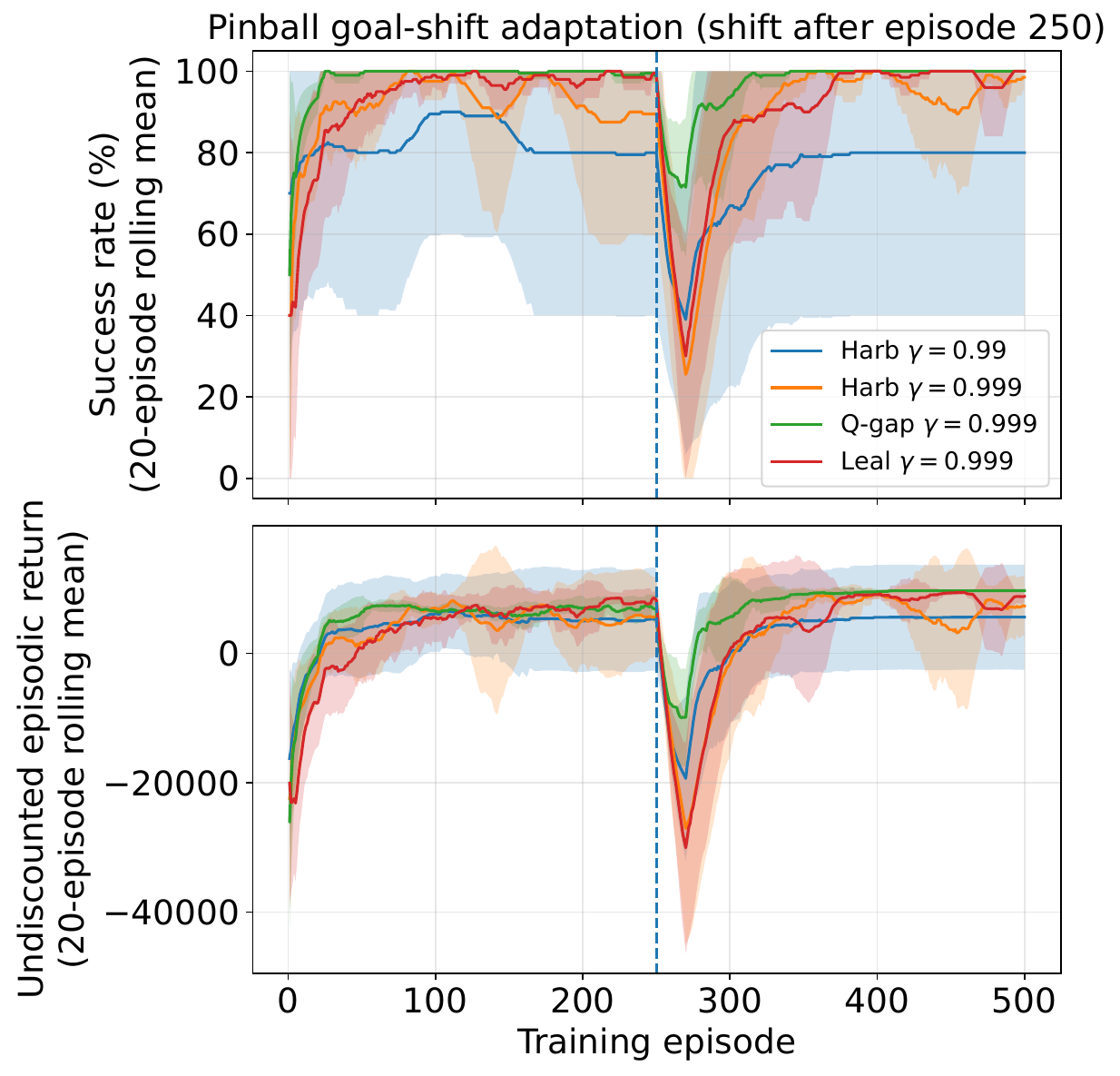}%
    }\hfill
    \subfloat[Altered Geometry]{%
        \includegraphics[width=0.5\textwidth]{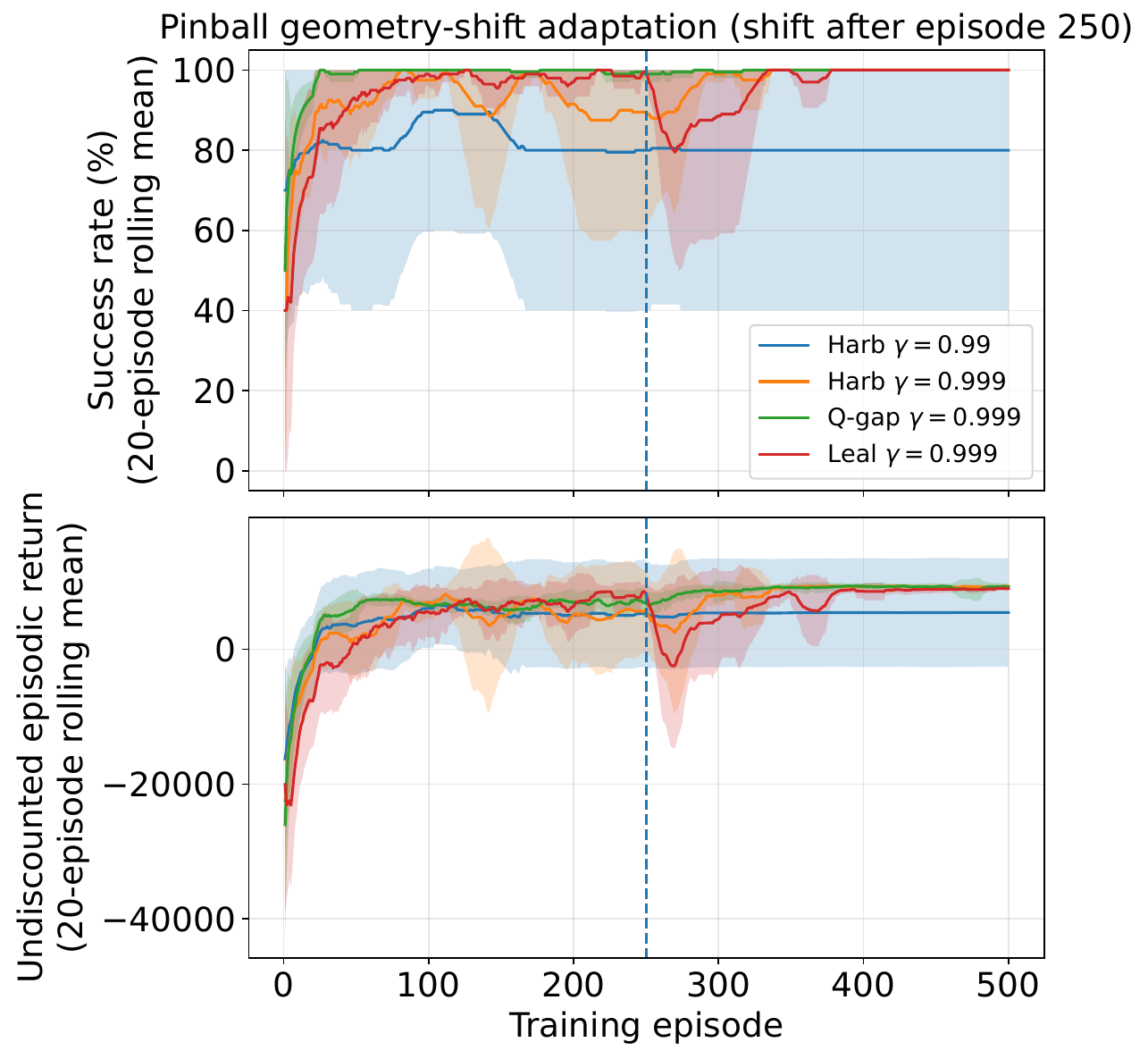}%
    }
    \caption{Pinball learning curves during training. The upper plots in
    each subplot show success rate over a 20-episode rolling window,
    while the lower plots show undiscounted episodic return. Curves
    represent means over ten seeds and shaded bands indicate $\pm$ one
    standard deviation. The environment changes after episode 250, so
    windows near the transition may contain episodes from both layouts.}
    \label{fig:adaptation}
\end{figure*}

% The target is now relocated after episode 250. Doing so, changes the which behaviors are useful for the task while preserving physical geometry.
Adaptation becomes critical when previously-successful behaviors are no longer optimal. To create such a scenario, the goal (red circle in Fig.~\ref{fig:layouts}) is relocated after episode 250, while the start (blue circle in Fig.~\ref{fig:layouts}) and the environment geometry (corridors' layout) remain unchanged.
Table~\ref{tab:adaptation} and Fig.~\ref{fig:adaptation} summarize the response. For $\gamma=0.999$, $OC^2$ achieved $52.0 \pm 27.4\%$ mean success over the first ten post-shift episodes; compared with the other baselines, the fixed cost got $18.0 \pm 38.2\%$ whereas the flexible cost got $15.0 \pm 26.8\%$. Mean success over all 250 post-shift episodes was $96.76 \pm 3.66\%$ for $OC^2$, $87.80 \pm 6.32\%$ for fixed cost and $89.28 \pm 10.99\%$ for flexible cost. Each of ten seeds recovered successfully under each of those three conditions. 
% These observations support faster recovery under the tested goal relocation.
After the goal relocation, the value of previously-successful options decreases, naturally reducing the opportunity cost of abandoning them. This weakens the persistence pressure and enables the agent to transition more rapidly toward alternative behaviors suited to the new objective. Therefore, the same mechanism that promotes persistence in stationary environments automatically reduces persistence after an environmental shift, eliminating the need for manually-tuned adaptation rules.

\subsection{Pinball: changed obstacle geometry}

% In this configuration, we changed the obstacle geometry, in which, we made sure it blocked the agent's learnt trajectory before the shift.
Unlike the goal-relocation scenario, this experiment changes the environment dynamics rather than the task objective. The obstacle geometry is modified so that the trajectory learned before the shift becomes inaccessible, forcing the agent to adapt its behavior.

$OC^2$ maintains $100.0 \pm 0.0\%$ success over the first ten post-shift episodes and $99.92 \pm 0.17\%$ over the entire run. In contrast, the flexible-cost baseline achieved $74.0 \pm 31.7\%$ and $96.12 \pm 7.29\%$, respectively, while, the fixed-cost baseline achieved $87.0 \pm 30.9\%$ and $98.80 \pm 2.32\%$, respectively. $OC^2$ also has a shorter mean post-change episodes: 198.3 compared to 759.7, 373.8 primitive steps, respectively. 

Unlike a fixed deliberation cost, the proposed mechanism does not enforce persistence uniformly. As the estimated utility of the active option changes following the geometry modification, the deliberation cost adapts accordingly, allowing persistence to remain aligned with expected future return.

% In conclusion, these 3 configurations showed how $OC^2$ learnt shorter path and was more persistent.
Taken together, these three configurations support the hypothesis that option persistence should be governed by the opportunity cost of abandoning the current behavior. The resulting $OC^2$ mechanism learned shorter paths while exhibiting greater persistence and adaptability.

\section{Discussion and Limitations}
\label{Discussion and Limitations}

$OC^2$ ties the persistence signal for an option to its relative state-option value function $Q_\Omega^C(s,\omega)$ with respect to competing alternatives. The same option may incur a high termination penalty in states where it remains valuable and a low penalty where superior alternatives exist. As the estimated advantage $A_\Omega^C(s,\omega)$ diminishes, the target penalty diminishes accordingly. Persistence therefore adapts across states according to the relative value of competing options, and evolves naturally in response to the agent's changing value estimates. This perspective also helps explain the improved adaptation observed after environmental changes. Because persistence pressure is tied to option values rather than fixed heuristics, the same mechanism that promotes commitment in stationary settings automatically encourages switching when the value landscape changes.

That said, this study still has the following limitations. First, the effectiveness of the proposed mechanism depends on the critic's ability to accurately rank competing options, since errors in option value estimation directly affect the learned persistence behavior. A second limitation concerns the adaptation speed of the cost predictor. Following environmental changes, outdated value-gap estimates may temporarily persist and induce persistence patterns that no longer reflect the current value ordering of the options. The learned cost also enters the critic target, and the convergence properties of these coupled updates remain an open question. Finally, our approach inherits the assumption that all options are available in every state. In practical settings with restricted initiation sets, the framework would need to be extended to predict option applicability and handle states in which no alternative options are available.

\section{Conclusion}
\label{conclusion}

We presented the Opportunity-Cost Option-Critic ($OC^2$) framework, which utilizes the positive gap between the current state-option value estimate versus its best distinct alternative to define an adaptive value-dependent deliberation cost target. The mechanism encourages persistence when the current option remains preferable and removes that pressure when a better alternative emerges. A cost-augmented formulation clarifies the actor updates and explains the need for the auxiliary cost-fitting objective. Tested in two environments and across ten random seeds, the $OC^2$ approach produced shorter Four Rooms trajectories, improved recovery following a Pinball goal relocation, and maintained high success rates under geometry alterations. 
% Together, these findings widens the horizon for further investigation of value-adaptive persistence.
Collectively, these results support the view that option persistence should be governed by the opportunity cost of abandoning the current behavior rather than by fixed heuristics or duration-dependent penalties, motivating further study of value-adaptive persistence in hierarchical reinforcement learning.

\end{document}